%% file: root.tex
\documentclass[letterpaper, 10 pt, conference]{ieeeconf}

\usepackage{bbm}
\usepackage{cite}
\usepackage{url}
\usepackage{comment}
\usepackage{algorithmic}
\usepackage{graphicx}
\usepackage{textcomp}
\usepackage{xcolor}
\usepackage{amsmath,amssymb,amsfonts}

\usepackage{multirow}

\usepackage{booktabs}

\usepackage{caption}

\usepackage{xcolor}

\usepackage[colorlinks,linkcolor=blue]{hyperref}

\include{shortcuts}

\include{acronyms}

\IEEEoverridecommandlockouts                              
\title{\LARGE \bf
Multi-Camera Unified Pre-training via 3D Scene Reconstruction
}

\author{Chen Min$^{1}$ and Liang Xiao$^{2}$ and Dawei Zhao$^{2,*}$ and Yiming Nie$^{2,*}$ and Bin Dai$^{2}$
\thanks{$^{1}$School of Computer Science, Peking University, 
	Beijing, China, 100871}%
\thanks{$^{2}$Unmanned Systems Research Center, NIIDT, 
		Beijing, China, 10073}%
\thanks{$^{*}$ Corresponding authors. Email: {\tt\small adamzdw@163.com, e\_ming\_e\_ming@163.com}}%
}

\begin{document}

\maketitle
\thispagestyle{empty}
\pagestyle{empty}

\begin{abstract}
Multi-camera 3D perception has emerged as a prominent research field in autonomous driving, offering a viable and cost-effective alternative to LiDAR-based solutions. The existing multi-camera algorithms primarily rely on monocular 2D pre-training. However, the monocular 2D pre-training overlooks the spatial and temporal correlations among the multi-camera system. To address this limitation, we propose the first multi-camera unified pre-training framework, called UniScene, which involves initially reconstructing the 3D scene as the foundational stage and subsequently fine-tuning the model on downstream tasks. Specifically, we employ Occupancy as the general representation for the 3D scene, enabling the model to grasp geometric priors of the surrounding world through pre-training. A significant benefit of UniScene is its capability to utilize a considerable volume of unlabeled image-LiDAR pairs for pre-training purposes. The proposed multi-camera unified pre-training framework demonstrates promising results in key tasks such as multi-camera 3D object detection and surrounding semantic scene completion. When compared to monocular pre-training methods on the nuScenes dataset, UniScene shows a significant improvement of about 2.0\% in mAP and 2.0\% in NDS for multi-camera 3D object detection, as well as a 3\% increase in mIoU for surrounding semantic scene completion. By adopting our unified pre-training method, a 25\% reduction in 3D training annotation costs can be achieved, offering significant practical value for the implementation of real-world autonomous driving. 
Codes are publicly available at \url{https://github.com/chaytonmin/UniScene}.
\end{abstract}

\section{Introduction}
\label{sec:intro}

The multi-camera 3D perception systems in autonomous driving offer a cost-effective solution to gather $360^\circ$ environmental information around vehicles, making it a hot research area recently~\cite{survey1,survey2,bevfusion,chen2023futr3d,li2022unifying}. However, current multi-camera 3D perception models~\cite{detr3d,bevformer,bevdet,bevdepth} usually rely on pre-trained ImageNet models~\cite{imagenet} or depth estimation models~\cite{detr3d} on monocular images. These models fail to take into account the inherent spatial and temporal correlations presented in multi-camera systems. Additionally, while monocular pre-training enhances the capability of image feature extraction, it does not address the pre-training requirements of subsequent tasks. 
\begin{figure}[t]
	\centering
	\includegraphics[width=0.5\textwidth]{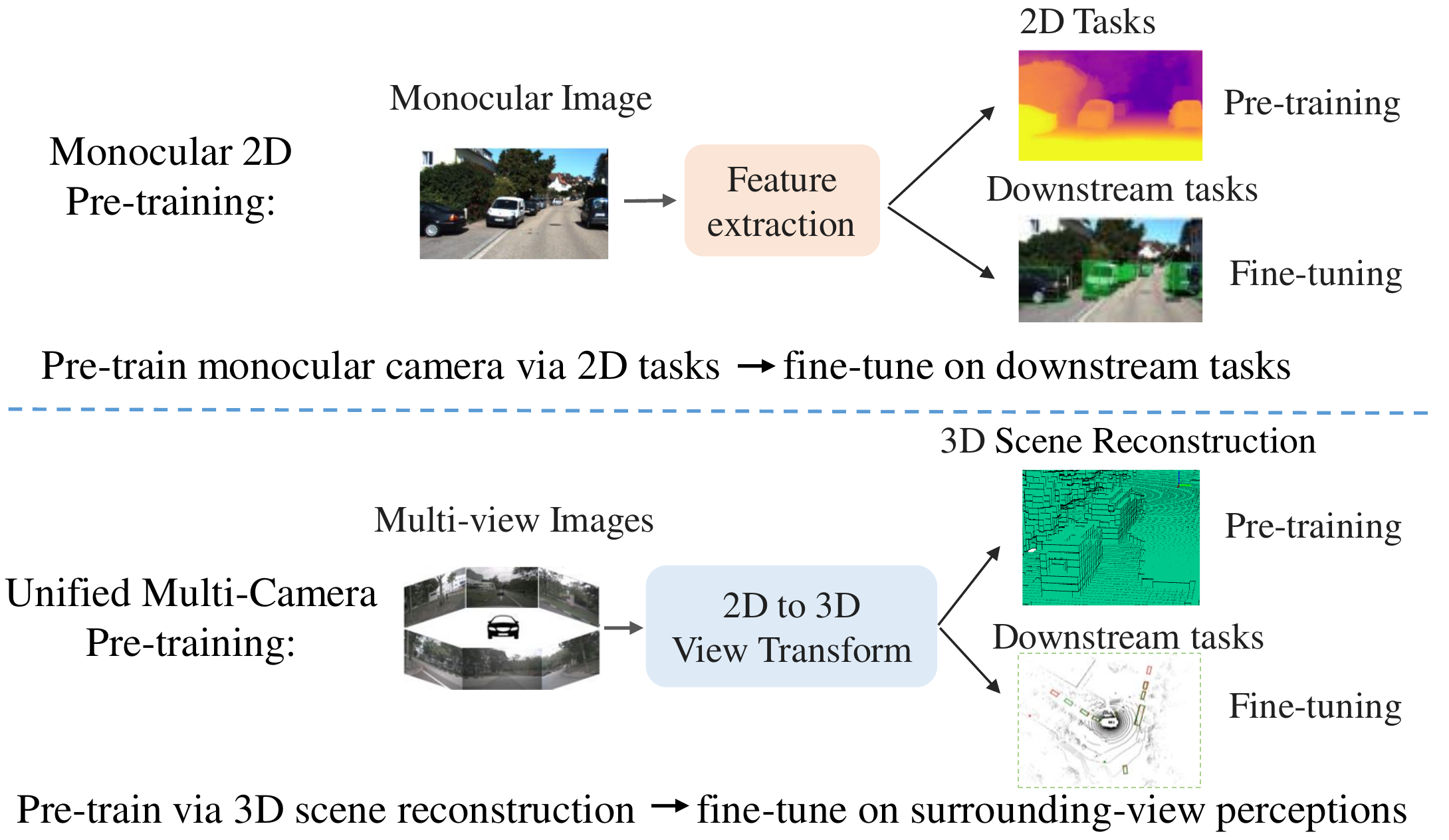} 
	\caption{Comparison between monocular pre-training and our unified multi-camera pre-training. Monocular pre-training only enhances the capability of the feature extraction from a single view, whereas our proposed multi-view unified pre-training enables the incorporation of temporal and spatial information from multi-view images through 3D scene reconstruction for pre-training.}
	\label{fig:compare1}
\end{figure}

Autonomous driving vehicles collect vast amounts of image-LiDAR pairs, which contain valuable 3D spatial and structural information. Thus, effectively utilizing these unlabeled image-LiDAR pairs can be beneficial for enhancing the performance of autonomous driving systems. Recent studies, such as BEVDepth~\cite{bevdepth} and DD3D~\cite{dd3d}, have underscored the significance of depth estimation in visual-based perception algorithms (Depth labels generated through LiDAR). Monocular depth estimation plays a crucial role in acquiring spatial position information for objects. However, depth estimation methods typically focus on estimating the depth of object surfaces, neglecting the holistic 3D structure of objects and occluded elements. For $360^\circ$ multi-camera perception systems, 3D occupancy stands as a comprehensive portrayal of the environment, offering a geometry-aware insight into the surroundings~\cite{occnet,occ_survey}. Achieving precise geometric occupancy prediction is instrumental in enhancing the overall 3D perception accuracy within multi-camera perception systems~\cite{occ_survey}. Hence, in the field of autonomous driving perception, as illustrated in Figure~\ref{fig:compare1}, the pre-training of models would yield greater benefits by prioritizing the reconstruction of the entire occupancy grids of the 3D scene, rather than solely emphasizing depth prediction. 

Humans possess the remarkable ability to mentally reconstruct the complete 3D geometry of occluded scenes, which is crucial for recognition and understanding~\cite{li2023voxformer}. To imbue the perception system of autonomous vehicles with a similar capability, we propose a multi-camera unified pre-training method, called UniScene. Our approach leverages the intuitive concept of using the multi-camera system to reconstruct the 3D scene as the foundational stage, followed by fine-tuning downstream tasks. In the case of multi-camera BEV perception, the input multi-camera images are transformed to the BEV space using advanced techniques like LSS~\cite{lss} or Transformer~\cite{detr3d}, and then a geometric occupancy prediction head is incorporated to learn the 3D occupancy distribution, thereby enhancing the model's understanding of the 3D surrounding scene. Due to the sparsity of single-frame point clouds, we employed multi-frame point cloud fusion as the ground truth for occupancy label generation. The decoder was solely used for pre-training, while the well-trained model was utilized to initialize the multi-camera perception models. By designing an effective multi-camera unified pre-training method, we enable the pre-trained model to exploit the rich spatial and temporal information inherent in the unlabeled data. This not only improves the model's ability to understand complex 3D scenes but also reduces reliance on costly and time-consuming manual 3D annotation.

To evaluate the effectiveness of our approach, we conducted extensive experiments using the widely used autonomous driving dataset nuScenes~\cite{nuscenes}. The experimental results demonstrate the superiority of our multi-camera unified pre-trained model compared to existing monocular pre-training methods across various 3D perception tasks, including 3D object detection and semantic scene completion.
In the 3D object detection task, the proposed UniScene achieves a significant improvement of 2.0\% in mAP and 2.0\% in NDS when compared to monocular pre-training methods. This indicates that our model is better equipped to accurately detect and localize objects in a 3D environment.
For the semantic scene completion task, UniScene demonstrates a noteworthy improvement of approximately 3\% in mIoU, indicating that our model is more effective in reconstructing and predicting the semantic labels of the surrounding environment.	
The superior performance of our model can be attributed to its ability to effectively leverage unlabeled data, as well as its consideration of spatial and temporal correlations. By incorporating information from multiple camera views, our model can better capture the rich contextual and temporal information present in the scene, leading to enhanced perception capabilities in autonomous driving scenarios. 

The main contributions of this work are listed below:
\begin{itemize}
	\item We define the task of multi-camera unified pre-training and propose the first unified pre-training framework. This framework utilizes Occupancy as a holistic representation of the 3D scene, allowing the model to acquire geometric insights about the surrounding world through pre-training.
	\item UniScene's pre-training process is label-free, enabling the utilization of massive amounts of image-LiDAR pairs collected by autonomous vehicles to build a Fundational Model.
	\item By adopting our unified pre-training method, a 25\% reduction in costly 3D annotation can be achieved, offering significant practical value for the implementation of real-world autonomous driving.
\end{itemize}

\begin{figure*}[t]
	\centering
	\includegraphics[width=1\textwidth]{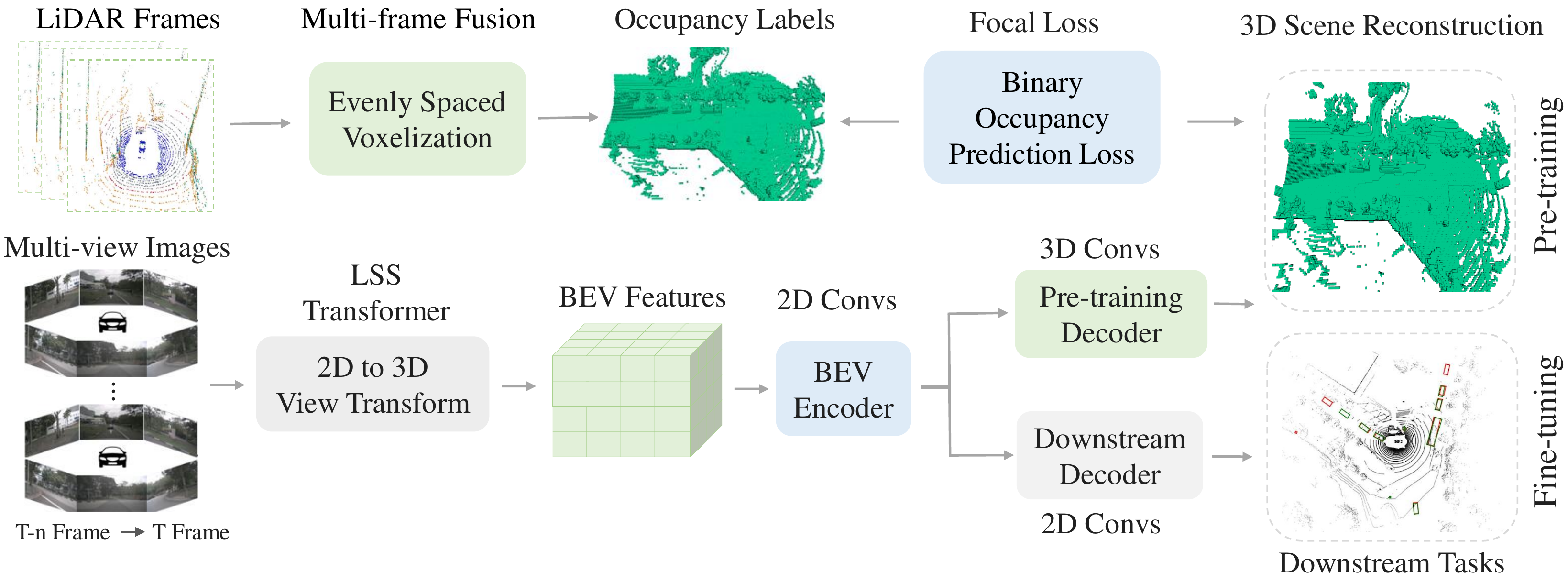}
	\caption{The overall architecture of UniScene. We employ Occupancy as the general representation for the 3D scene. Through pre-training on a substantial dataset of image-LiDAR pairs, our multi-camera perception model acquires prior knowledge about surrounding 3D scene distributions. Subsequently, we fine-tune the model for specific downstream tasks. The labels for occupancy are generated by fusing data from multiple frames of LiDAR point clouds. After pre-training, the lightweight decoder is discarded, and the encoder is used to warm up the backbones of downstream tasks.
	}
	\label{fig:flowchart}
\end{figure*}

\section{Related Work}
\subsection{Multi-Camera 3D Perception}

In the field of autonomous driving, vision-based 3D perception conducted in bird's eye view has gained significant attention in recent years~\cite{zhang2022beverse,videobev}. Learning-based BEV perception methods, based on 2D-to-3D view transformation, can be broadly categorized into geometry-based and Transformer-based approaches~\cite{sts,solofusion}. One of the early geometry-based methods is LSS \cite{lss}, which lifts each individual image into a frustum of features for each camera and then combines them into a rasterized BEV grid. Building upon LSS, BEVDet \cite{bevdet} introduces image-view and BEV space data augmentation techniques. 
BEVDepth \cite{bevdepth} demonstrates the significance of depth and improves the quality of BEV features by incorporating explicit depth supervision from LiDAR. 
BEVStereo \cite{bevstereo} and STS \cite{sts} leverage temporal multi-view stereo methods to enhance depth precision. SOLOFusion~\cite{solofusion} and VideoBEV~\cite{videobev} explore long-term temporal fusion for multi-view 3D perception. 
DETR3D \cite{detr3d} is the first Transformer-based BEV method, which defines object queries in 3D space and learns from multi-view image features using a transformer decoder. Building upon DETR3D, PETR \cite{petr} enhances the approach with position embedding transformations, while BEVFormer \cite{bevformer} introduces temporal self-attention to fuse historical BEV features. 
UniAD \cite{uniad} extends BEVFormer to enable multi-task learning in BEV space. 
Although the existing BEV perception methods have shown promising performance, they are typically initialized with ImageNet pre-trained models \cite{imagenet} or depth pre-trained models \cite{dd3d} trained on monocular images. However, there is a lack of unified pre-training methods that effectively leverage the geometric structure of multi-camera inputs.

\subsection{Lable-free Pre-training}

Lable-free Pre-training has gained significant popularity in recent years as it eliminates the need for expensive data annotation. 
DeepCluster \cite{deepcluster} and SwAV \cite{swav} leverage k-means clustering to obtain pseudo-labels, which are then used to train the network. Moco \cite{moco} and BYOL \cite{byol} construct contrastive views for self-supervised learning. Additionally, methods like MAE \cite{mae} and BEiT \cite{beit} employ a random patch masking approach where missing pixels or features are reconstructed using a simple autoencoder framework. Voxel-MAE \cite{voxel-mae}, ALSO~\cite{also}, and ISCC~\cite{iscc} propose predicting occupancy for LiDAR perception as the pretext task. In this work, we introduce the first multi-camera unified pre-training approach that utilizes $360^\circ$ geometric occupancy prediction for vision-based perception methods.
\section{Methodology}

This section elaborates on the network architecture of UniScene, as shown in Figure~\ref{fig:flowchart}. First, the vision-based BEV perception methods are reviewed in Section~\ref{review}. Then, the proposed geometric occupancy pre-training is introduced in Section~\ref{geo}, along with the comparison with existing methods in Section~\ref{com} and destription of occupancy label generation in Section~\ref{label}. 

\subsection{Review of BEV Perception}
\label{review}

As described in related works, there are two main learning-based methods to convert 2D images to 3D space: LSS-based~\cite{lss} and Transformer-based~\cite{detr3d} view transformation. Our method is not limited to specific view transformation methods. In the following sections, we will provide an overview of the workflow of multi-camera perception algorithms based on the bird's eye view.

The multi-camera input images, denoted as $I=\{I_i, i=1, 2, ..., N_{view}\}$, are initially processed by an image backbone network (e.g., ResNet-101~\cite{resnet}), generating feature maps  $F_{2d}=\{F_{2d}^{i}\}_{i=1}^{N_{view}}$ for each camera view. These features are then fed into a 2D-to-3D view transformation operation to project them onto a unified bird's eye view representation, denoted as $F_{bev}\in \mathbb{R}^{C\times H\times W}$. By incorporating specific heads, diverse autonomous driving perception tasks can be accomplished on the bird's eye view, including 3D object detection, map segmentation, object tracking, and more~\cite{uniad}. Current BEV perception algorithms~\cite{bevformer,detr3d,bevdet} predominantly rely on feature extraction models (e.g., ImageNet~\cite{imagenet}) or depth estimation models (e.g., V2-99~\cite{dd3d}) trained on monocular images. However, these approaches fail to consider the interplay and correlation between images captured from different camera views and frames. Consequently, there is a lack of a multi-camera unified pre-training model. In order to fully exploit the spatial and temporal relationships between different camera views, we propose a multi-camera unified pre-training model.

\subsection{Multi-camera Unified Pre-training}
\label{geo}

Methods such as BEVDepth~\cite{bevdepth} and DD3D~\cite{dd3d} demonstrate the importance of depth estimation for visual-based perception algorithms. However, depth estimation can only estimate the position of the object's surface, ignoring the occlusions of objects. For multi-camera systems, precision 3D occupancy grid prediction is beneficial to the accuracy of perception. 

\subsubsection{Geometric Occupancy Decoder}	
To predict 3D geometric occupancy from BEV features $F_{bev}$, we initially transform the BEV features into $F_{bev}^{'}\in \mathbb{R}^{C^{'}\times D\times H\times W}$, where D represents the number of height channels, and $C=C^{'}\times D$. Subsequently, we employ a 3D decoder that is specifically designed to generate 3D geometric occupancy. Our decoder comprises lightweight 3D convolution layers, with the final layer providing the probability of each voxel containing points. The decoder's output is denoted as $\textbf{P}\in \mathbb{R}^{D\times H \times W \times 1}$. During pre-training, the decoder's primary purpose is to reconstruct occupied voxels.

\subsubsection{Pre-training Target}
Taking into account the sparse nature of single-frame LiDAR point clouds and the potential inaccuracies arising from fusing a large number of frames due to the presence of dynamic objects, we fuse the LiDAR point clouds from some keyframes for the occupancy labels generation. In line with the standard practice in 3D perception models~\cite{second,pv_rcnn,centerpoint,cylinder3d}, the LiDAR point clouds are partitioned into evenly spaced voxels. For the dimensions of the LiDAR point clouds along $Z\times Y\times X$ as $D\times H\times W$ respectively, the voxel size is determined as $v_Z\times v_H\times v_W$ correspondingly. 
The  occupancy of the voxels, i.e., containing points or not in each voxel, serve as the ground truth $\textbf{T} \in \{0,1\}^{D \times H\times W\times 1}$. 1 means occupied and 0 means free.

We introduce the binary geometric occupancy classification task for pre-training multi-camera perception models. The objective of this task is to train the network to accurately predict the geometric occupancy distribution of the 3D scene based on multi-view images. Taking into account the substantial number of empty voxels, predicting occupancy grids poses an imbalanced binary classification problem. To accomplish this, we compute the focal loss for binary occupancy classification, utilizing the predicted occupied values $\textbf{P}$ and the ground truth occupied voxels $\textbf{T}$:

\begin{equation} \label{loss}
loss = -\frac{1}{batch}\frac{1}{n}\sum_{i=1}^{batch}\sum_{j=1}^{n}\alpha_t \left(1 - \textbf{P}_t^{ij}\right)^\gamma \log(\textbf{P}_t^{ij}),
\end{equation}
where $\textbf{P}^{ij}$ is the predicted probability of voxel $j$ in the $i$-th
training sample. $n=D\times H \times W$ is the total number of voxels and batch is the number of batch sizes. The weighting factor $\alpha$ for positive/negative examples is set as 2 and the weighting factor $\gamma$ for easy/hard examples is 0.25. $\alpha_t=\alpha$ and $\textbf{P}_t^{ij} = \textbf{P}^{ij}$ for class 1. $\alpha_t=1-\alpha$ and $\textbf{P}_t^{ij} = 1-\textbf{P}^{ij}$ for class 0. 

\subsubsection{Pre-training for Surrounding Semantic Occupancy Prediction}

Recently, several algorithms such as TPVFormer~\cite{tpvformer}, OpenOccupancy~\cite{openoccupancy} and Occ3D~\cite{occ3d} have extended multi-camera BEV perception to the task of surrounding semantic scene completion~\cite{gan2023simple}. However, directly predicting the 3D semantics of multi-view images requires a large amount of 3D semantic annotation for training, which can be costly and time-consuming. To address this challenge, we propose to extend our multi-camera unified pre-training algorithm to the surrounding semantic scene completion task, that is, firstly perform geometric occupancy prediction, and then fine-tune on semantic scenes completion task.

\subsubsection{Semantic Occupancy Supervision}
The aforementioned geometry occupancy pre-training can leverage a substantial amount of unannotated image-LiDAR pairs for learning. If annotated data is available, explicit supervision for semantic occupancy can be incorporated during the training of the multi-camera perception model. By utilizing 3D annotations, it becomes possible to segment moving and static targets, thereby obtaining more precise occupancy grid ground truth that is fused from multiple LiDAR frames. This enables the model to learn from more accurate and reliable information, enhancing its perception capabilities in handling various scenarios.

\subsection{Comparision with Existing Methods}
\label{com}
\subsubsection{Comparision with Monocular Pre-training}

Currently, multi-camera perception algorithms employ either monocular image pre-training on ImageNet~\cite{imagenet} or depth estimation pre-training~\cite{dd3d}. As shown in Figure~\ref{fig:compare1}, our proposed multi-camera unified pre-training model offers several advantages over monocular pre-training:
(1) \textbf{Spatial-Temporal Integration}: By leveraging the spatial and temporal information from multiple camera views, the model can better comprehend the dynamic nature of the environment and make more accurate predictions.
(2) \textbf{Unified Representation}: The unified pre-training approach allows the model to learn a shared representation across different camera views, promoting better knowledge transfer and reducing the need for task-specific pre-training.
(3) \textbf{Perception of occluded areas}: Monocular depth estimation can only predict the surface positions of objects, while the proposed multi-camera unified pre-training method enables the overall 3D reconstruction of occluded objects.

\subsubsection{Comparision with Knowledge Distillation}
Recently, there have been advancements in knowledge distillation algorithms such as BEVDistill~\cite{bevdistill}, TiG-BEV~\cite{tigbev} and GeoMIM~\cite{geomim}, which aim to transfer knowledge from well-established 3D LiDAR models like CenterPoint~\cite{centerpoint} to multi-camera object detection algorithms. Similarly, our approach aims to leverage the rich spatial information presented in 3D point clouds and transfer it to multi-camera algorithms. Our unique pre-training algorithm eliminates the need for annotations or pre-trained LiDAR detection models, significantly reducing the 3D annotation requirements.

\subsubsection{Comparision with Surface Occupancy Reconstruction}
While previous approaches, such as Voxel-MAE~\cite{voxel-mae}, ALSO~\cite{also}, and ISCC~\cite{imagenet}, have predominantly concentrated on surface occupancy reconstruction for LiDAR pre-training, multi-camera perception faces a distinct challenge due to its inherent lack of 3D spatial geometric information when compared to LiDAR. In response to this limitation, our proposed method seeks to extend the scope by not only predicting surface details but also addressing occluded areas, thus enabling the reconstruction of a comprehensive 3D scene. Additionally, our multi-camera unified pre-training algorithm introduces an innovative spatio-temporal fusion mechanism, designed specifically to amalgamate both spatial and temporal data streams from multi-camera systems. This approach is tailored to enhance the effectiveness of pre-training for multi-camera perception.

\subsection{Occupancy Lable Generation}
\label{label}
Currently, the geometry occupancy labels used in the algorithm are obtained from LiDAR point clouds. In the future, it is also feasible to utilize point clouds generated from 3D scene reconstructions using techniques such as NeRF~\cite{nerf,nerf2} or MVS~\cite{mvs,mvsnet,aa-rmvsnet,bi}. However, the current Nerf-based and MVS-based 3D scene reconstruction still faces challenges such as insufficient accuracy and inability to handle dynamic objects. As Tesla plans to incorporate Nerf in the future~\cite{tesla}, we aspire to extend our method to incorporate Nerf's capabilities of scene reconstruction. 

\section{Experiments}

\begin{table*}[t]
	\centering
	\renewcommand{\arraystretch}{1.5} 
	\caption{Quantitative multi-camera 3D object detection performance on the nuScenes validation set.}
	\resizebox{\textwidth}{!}
	{
		{
			\begin{tabular}{c|c|c|c|c|c|c|c|c|c|c|c|c}
				\toprule
				\textbf{Type}&\textbf{Method}&\textbf{Pre-train}&\textbf{Backbone}&\textbf{Image Size}&\textbf{CBGS} &\textbf{mAP}$\uparrow$&\textbf{NDS}$\uparrow$&\textbf{mATE}$\downarrow$&\textbf{mASE}$\downarrow$&\textbf{mAOE}$\downarrow$&\textbf{mAVE}$\downarrow$&\textbf{mAAE}$\downarrow$ \\
				\midrule
				\multirow{4}*{Transformer~\cite{detr3d}}	
				&DETR3D~\cite{detr3d}&FCOS3D~\cite{fcos3d}&R101-DCN&900$\times$1600&$\checkmark$ &0.349&0.434&0.716&0.268&0.379&0.842&0.200\\
				\cline{2-13}
				&DETR3D~\cite{detr3d}&UniScene&R101-DCN&900$\times$1600&$\checkmark$ &$\textbf{0.360}^{\textcolor{teal} {+1.1\%}}$&$\textbf{0.461}^{\textcolor{teal} {+2.7\%}}$&\textbf{0.701}&\textbf{0.260}&\textbf{0.372}&\textbf{0.730}&\textbf{0.188}\\
				\cline{2-13}
				
				&BEVFormer~\cite{bevformer}&FCOS3D~\cite{fcos3d}&R101-DCN&900$\times$1600&$\times$ &0.416&0.517&0.673&0.274&0.372&0.394&0.198\\
				\cline{2-13}
				&BEVFormer~\cite{bevformer}&UniScene&R101-DCN&900$\times$1600&$\times$ &$\textbf{0.438}^{\textcolor{teal} {+2.2\%}}$&$\textbf{0.534}^{\textcolor{teal} {+1.7\%}}$&\textbf{0.656}&\textbf{0.271}&\textbf{0.371}&\textbf{0.348}&\textbf{0.183}\\
				
				\midrule 
				\midrule
				\multirow{4}*{LSS~\cite{lss}}&BEVDet~\cite{bevdet}&ImageNet~\cite{imagenet}&R-50&256$\times$704&$\times$&0.286&0.372& 0.724 &0.278 &0.590 &0.873 &0.247\\
				\cline{2-13}
				&BEVDet~\cite{bevdet}&UniScene&R-50&256$\times$704&$\times$&$\textbf{0.310}^{\textcolor{teal} {+2.4\%}}$&$\textbf{0.395}^{\textcolor{teal} {+2.3\%}}$& \textbf{0.701} &\textbf{0.259} &\textbf{0.578} &\textbf{0.852} &\textbf{0.230}\\
				\cline{2-13}
				
				&BEVDepth~\cite{bevdepth}&ImageNet~\cite{imagenet}&R-50&256$\times$704&$\times$ &0.351&0.475&0.639 &0.267 &0.479 &0.428 &0.198\\
				\cline{2-13}
				&BEVDepth~\cite{bevdepth}&UniScene&R-50&256$\times$704&$\times$ &$\textbf{0.376}^{\textcolor{teal} {+2.5\%}}$&$\textbf{0.492}^{\textcolor{teal} {+1.7\%}}$&\textbf{0.620}&\textbf{0.259}&\textbf{0.466}&\textbf{0.425}&\textbf{0.187}\\	
				\bottomrule
			\end{tabular}
	}}
	\label{tab:nuscenes_val}
\end{table*}

\begin{table*}[t]
	\centering
	\renewcommand{\arraystretch}{1.5}
	\caption{Quantitative multi-camera 3D object detection performance on the nuScenes test set.}
	\resizebox{\textwidth}{!}
	{
		{
			\begin{tabular}{c|c|c|c|c|c|c|c|c|c|c|c}
				\toprule
				\textbf{Method}&\textbf{Pre-train}&\textbf{Backbone}&\textbf{Image Size}&\textbf{CBGS} &\textbf{mAP}$\uparrow$&\textbf{NDS}$\uparrow$&\textbf{mATE}$\downarrow$&\textbf{mASE}$\downarrow$&\textbf{mAOE}$\downarrow$&\textbf{mAVE}$\downarrow$&\textbf{mAAE}$\downarrow$ \\
				\midrule
				\multirow{2}*{ DETR3D~\cite{detr3d}}&DD3D~\cite{dd3d}&v2-99&900$\times$1600&$\checkmark$&0.412&0.479&0.641 &\textbf{0.255} &\textbf{0.394} &0.845 &0.133\\
				\cline{2-12}
				&UniScene&v2-99&900$\times$1600&$\checkmark$&$\textbf{0.431}^{\textcolor{teal} {+1.9\%}}$&$\textbf{0.496}^{\textcolor{teal} {+1.7\%}}$&\textbf{0.621}&0.257&0.407&\textbf{0.783}&\textbf{0.123}\\
				\bottomrule
			\end{tabular}
	}}
	\label{tab:nuscenes_test}
\end{table*}

\begin{table*}[t]
	\centering
	\renewcommand{\arraystretch}{1.5}
	\caption{Quantitative segmentation performance on the 3D occupancy prediction challenge~\cite{occ}.}
	\resizebox{\textwidth}{!}{
		{
			\begin{tabular}{c|c|c|c|c|c|c|c|c|c|c|c|c|c|c|c|c|c|c|c|c|c}
				\toprule
				\textbf{Methods} &\textbf{Pre-train}&\textbf{Backbone}&\textbf{Image Size}&\textbf{mIoU}$\uparrow$ &\textbf{\rotatebox{90}{others}}&\textbf{\rotatebox{90}{barrier}}&\textbf{\rotatebox{90}{bicycle}}&\textbf{\rotatebox{90}{bus}}&\textbf{\rotatebox{90}{car}}&\textbf{\rotatebox{90}{construction}}&\textbf{\rotatebox{90}{motorcycle}}&\textbf{\rotatebox{90}{pedestrian}}&\textbf{\rotatebox{90}{traffic-cone}}&\textbf{\rotatebox{90}{trailer}}&\textbf{\rotatebox{90}{truck}}&\textbf{\rotatebox{90}{driveable}}&\textbf{\rotatebox{90}{others}}&\textbf{\rotatebox{90}{sidewalk}}&\textbf{\rotatebox{90}{terrain}}&\textbf{\rotatebox{90}{manmade}}&\textbf{\rotatebox{90}{vegetation}}\\
				\midrule
				BEVFormer~\cite{bevformer} &FCOS3D~\cite{fcos3d} &R101-DCN~\cite{resnet}&900$\times$1600&23.70&	10.24&	36.77&	11.70&	29.87&	38.92&	10.29&	22.05&	16.21&	14.69&	27.44&	23.13&	48.19&	33.10&	29.80&	17.64&	19.01&	13.75\\
				\midrule
				BEVStereo~\cite{bevstereo}&BEVDet4D~\cite{bevdet} &Swin-B~\cite{swin}&512$\times$1408&42.78&	22.45&	47.95&	28.13&	40.29&	53.79&	27.60&	35.18&	29.64&	31.69&	45.49&	37.71&	81.88&	49.16&	55.03&	51.00&	50.87&	39.44\\
				\midrule
				BEVStereo~\cite{bevstereo}&UniScene &Swin-B~\cite{swin}&512$\times$1408&$\textbf{45.92}^{\textcolor{teal} {+3.14\%}}$&\textbf{26.21}&	\textbf{53.06}&	\textbf{33.41}&	\textbf{42.77}&	\textbf{56.57}&	\textbf{28.99}&	\textbf{39.92}&	\textbf{32.31}&	\textbf{34.89}&	\textbf{49.59}&	\textbf{40.28}&	\textbf{82.88}&	\textbf{52.29}&	\textbf{57.77}&	\textbf{53.58}&	\textbf{53.94}&	\textbf{42.25}
				\\
				\bottomrule
			\end{tabular}
	}}
	\label{tab:nuscenes_seg}
\end{table*}

\subsection{Experimental Setup}
	\paragraph{Dataset}
	The nuScenes dataset~\cite{nuscenes} includes a comprehensive 360-degree panoramic perception system utilizing six cameras, while Waymo~\cite{waymo} employs five cameras that do not provide complete 360-degree coverage and KITTI~\cite{kitti} offers a forward field of view only.
	Thus, we only conducted extensive experiments on the nuScenes dataset. The nuScenes dataset consists of 28,130 training samples and 6,019 validation samples. 
	To evaluate the performance of our method, we utilized the official evaluation metrics provided by nuScenes, including the nuScenes Detection Score (NDS), mean average precision (mAP), average translation error (ATE), average scale error (ASE), average orientation error (AOE), average velocity error (AVE), and average attribute error (AAE). These metrics provide a comprehensive assessment of the detection, localization, and attribute prediction capabilities of our approach. 
	\begin{minipage}[t]{.5\linewidth} 
		\includegraphics[width=.99\linewidth]{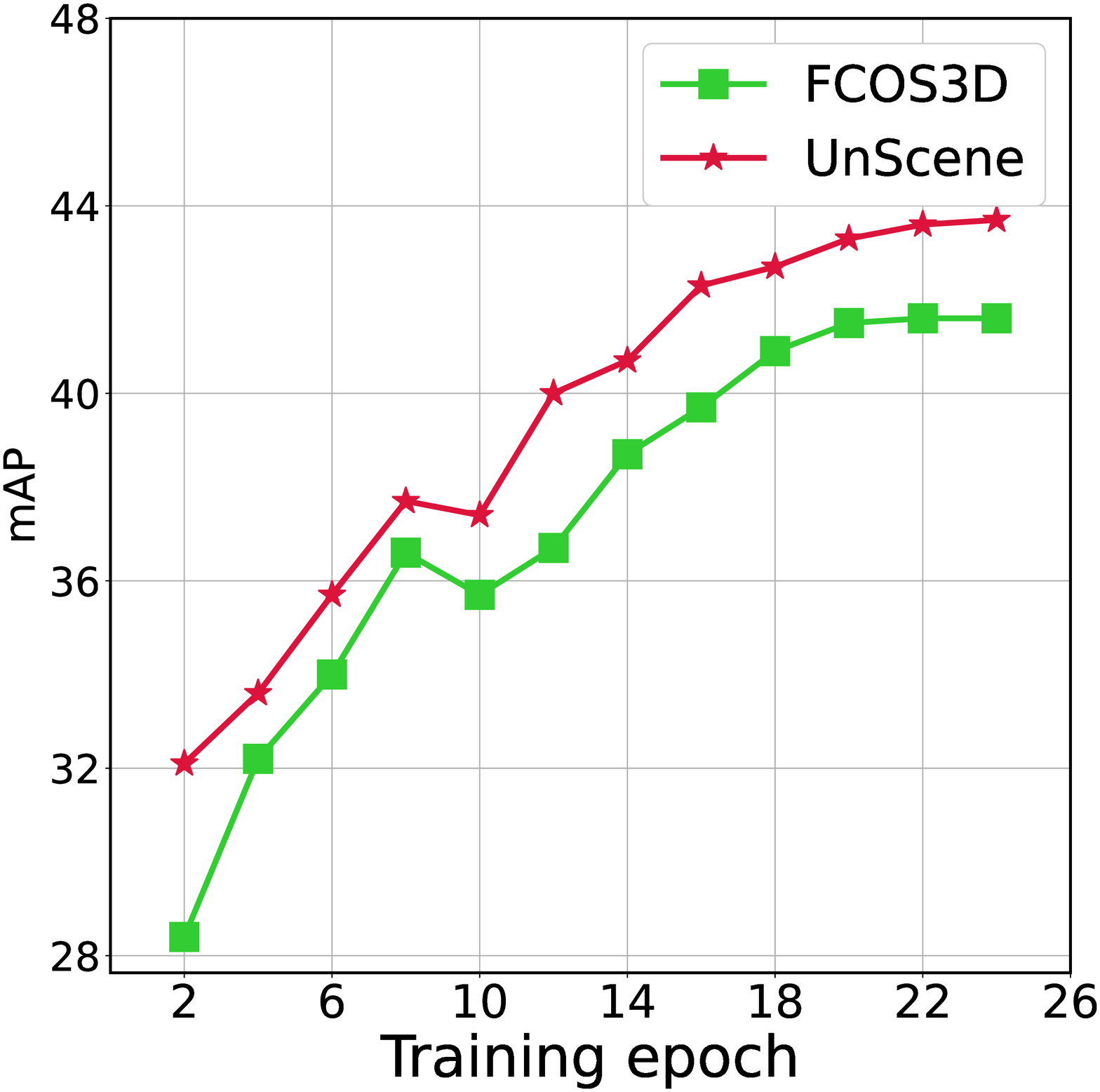}
		\captionof{figure}{Performance curves.
		}
		\label{fig:epoch}
	\end{minipage}%
	\begin{minipage}[t]{.5\linewidth} 
		\includegraphics[width=.99\linewidth]{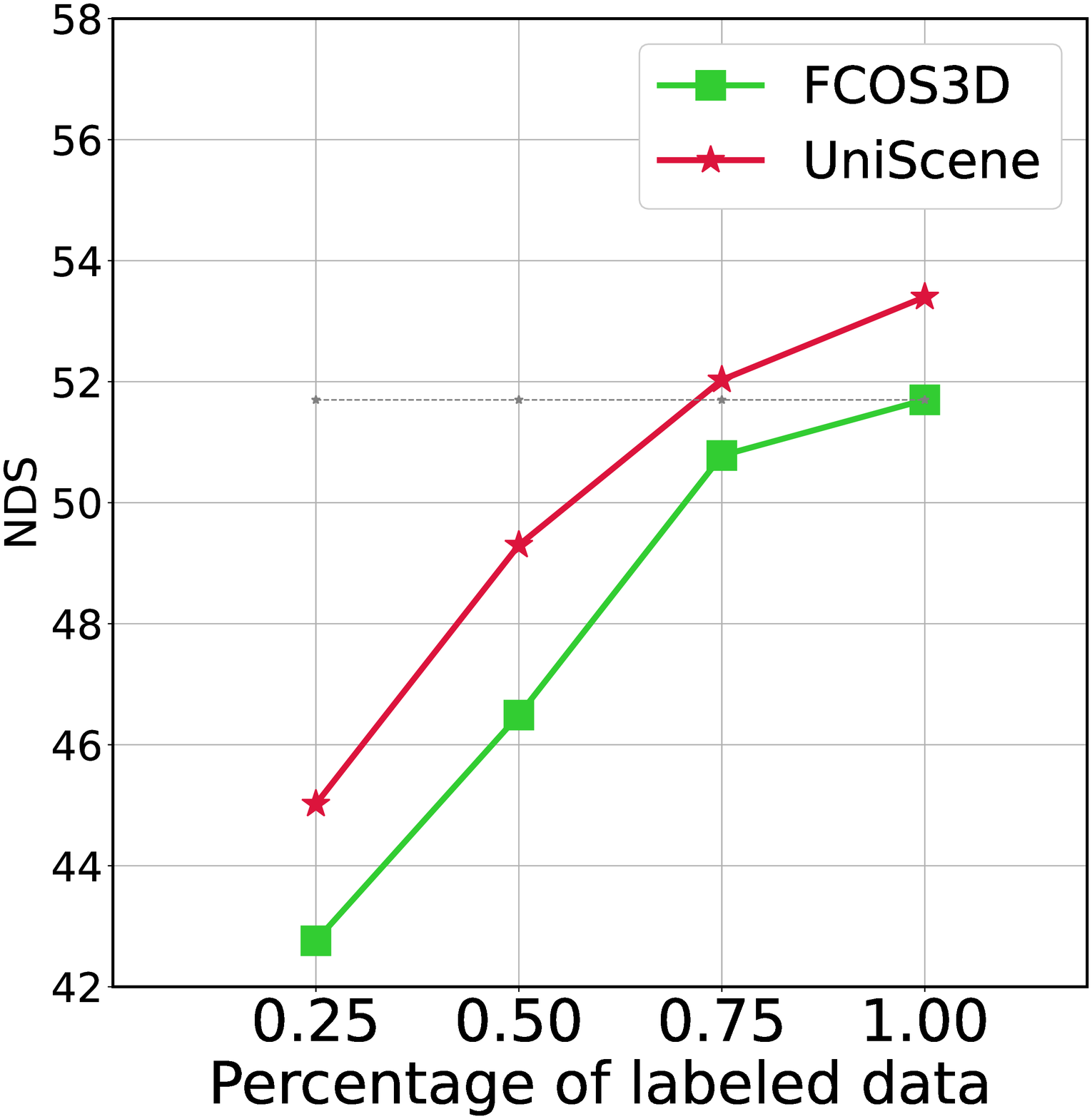}
		\captionof{figure}{Label-efficiency.
		}
		\label{fig:efficient}
	\end{minipage}%
	
	\paragraph{Implementation Details}
	
We adopted the training settings from the existing methods: DETR3D~\cite{detr3d} and BEVFormer~\cite{bevformer}, which are two Transformer-based methods, and BEVDet~\cite{bevdet}, BEVDepth~\cite{bevdepth}, and BEVStereo~\cite{bevstereo}, which are three LSS-based methods. The voxel size is set as $16\times200\times200$ for DETR3D and BEVFormer, and $16\times128\times128$ for BEVDet, BEVDepth and BEVStereo. We performed pre-training for a total of 24 epochs. The occupancy decoder consists of two layers of 3D convolutional layers. For more detailed information about the parameter setups, please refer to the papers of DETR3D, BEVFormer, BEVDet, BEVDepth and BEVStereo. We first pre-train UniScene and then use it to initialize the model. During fine-tuning, we do not freeze any modules, keeping the training settings consistent with the original ones.
All experiments were conducted using 8 Nvidia Tesla A40 GPU cards.
\subsection{Results on Downstream Tasks}

\subsubsection{3D Object Detection}

\begin{table}[t]
	\centering
	\caption{Comparison with BEVDistill~\cite{bevdistill}.}
	\setlength{\tabcolsep}{7.9mm}
	{
		\begin{tabular}{c|c|c}
			\toprule
			\textbf{Methods} &\textbf{mAP}$\uparrow$&\textbf{NDS}$\uparrow$\\
			\midrule
			BEVFormer~\cite{bevformer}&0.352&0.423 \\
			BEVDistill~\cite{bevdistill}&0.386&0.457\\
			\midrule
			UniScene&\textbf{0.389}&\textbf{0.459}\\
			\bottomrule
	\end{tabular}}
	\label{tab:kd}
\end{table}

We first conducted an evaluation of UniScene on the validation set of nuScenes. As shown in Table~\ref{tab:nuscenes_val}, our multi-camera unified pre-training method exhibited significant improvements over monocular FCOS3D~\cite{fcos3d}. It surpassed DETR3D~\cite{detr3d} by achieving a 2.7\% increase in NDS and 1.1\% in mAP. Additionally, it outperformed BEVFormer~\cite{bevformer} with a 1.7\% improvement in NDS and 2.2\% in mAP. We present the convergence curve of BEVFormer~\cite{bevformer} in Figure~\ref{fig:epoch}. Our unified pre-training significantly enhances BEVFormer~\cite{bevformer} at the initial epoch, achieving a 4\% increase in NDS. This demonstrates that our unified pre-training method delivers accurate object position information from a global perspective. 
Similarly, for the LSS-based methods BEVDet~\cite{bevdet} and BEVDepth~\cite{bevdepth}, initialized with monocular pre-training on ImageNet~\cite{imagenet}, our multi-camera unified pre-training method showed an improvement of approximately 2.0\% in NDS and 2.0\% in mAP.  

For further validation, we conducted additional experiments on the nuScenes test set to validate the effectiveness of our proposed multi-camera unified pre-training method via 3D scene reconstruction compared to pre-training based on monocular depth estimation. The submission time for test set of nuScenes is limited to 3. DETR3D provides the complete training code to reproduce the test results on the leaderboard. Hence, to ensure a fair comparison of pre-trained models, we have exclusively presented results for DETR3D on test set. As presented in Table~\ref{tab:nuscenes_test}, our multi-camera unified pre-training method demonstrated a significant improvement of about 1.8\% in both mAP and NDS compared to the DETR3D~\cite{detr3d} pre-trained on DD3D~\cite{dd3d} for depth estimation. This highlights the effectiveness and superiority of our pre-training approach in enhancing the performance of 3D perception tasks. Compared to the monocular depth estimation approach of DD3D~\cite{dd3d}, our pre-training method considers the complete 3D structure of objects, beyond the partial surfaces captured by LiDAR.
Moreover, it incorporates the learning of multi-view and temporal information, allowing for a more comprehensive understanding of the scene. 
The above results indicated that our proposed UniScene model has a promising application in autonomous driving.	

We also compared our proposed multi-camera unified pre-training method with the knowledge distillation approach BEVDistill~\cite{bevdistill}. As shown in Table~\ref{tab:kd}, our method demonstrates comparable performance to the knowledge distillation method trained on annotated LiDAR point clouds data. It is worth noting that our approach offers higher efficiency and broader applicability since it does not rely on data annotation or training of LiDAR point clouds models as BEVDistill~\cite{bevdistill}.

\subsubsection{Semantic Occupancy Prediction}

We also evaluated the performance of our proposed multi-camera unified pre-training method on the task of multi-camera semantic scene completion. Compared to BEV perception, the task of predicting semantic labels for each voxel in 3D space, known as surrounding semantic scene completion, is more challenging. To tackle this challenge, we decomposed the task into two steps: first reconstructing the 3D scene as the fundamental model and then simultaneously reconstructing and predicting semantics. As shown in Table~\ref{tab:nuscenes_seg}, on the 3D occupancy prediction challenge~\cite{occ}, our algorithm achieved a 3\% improvement in mIoU compared to BEVStereo~\cite{bevstereo}, highlighting the effectiveness of UniScene in addressing the complexities of surrounding semantic occupancy prediction.

\begin{figure*}[t]
	\centering
	\includegraphics[width=0.98\textwidth]{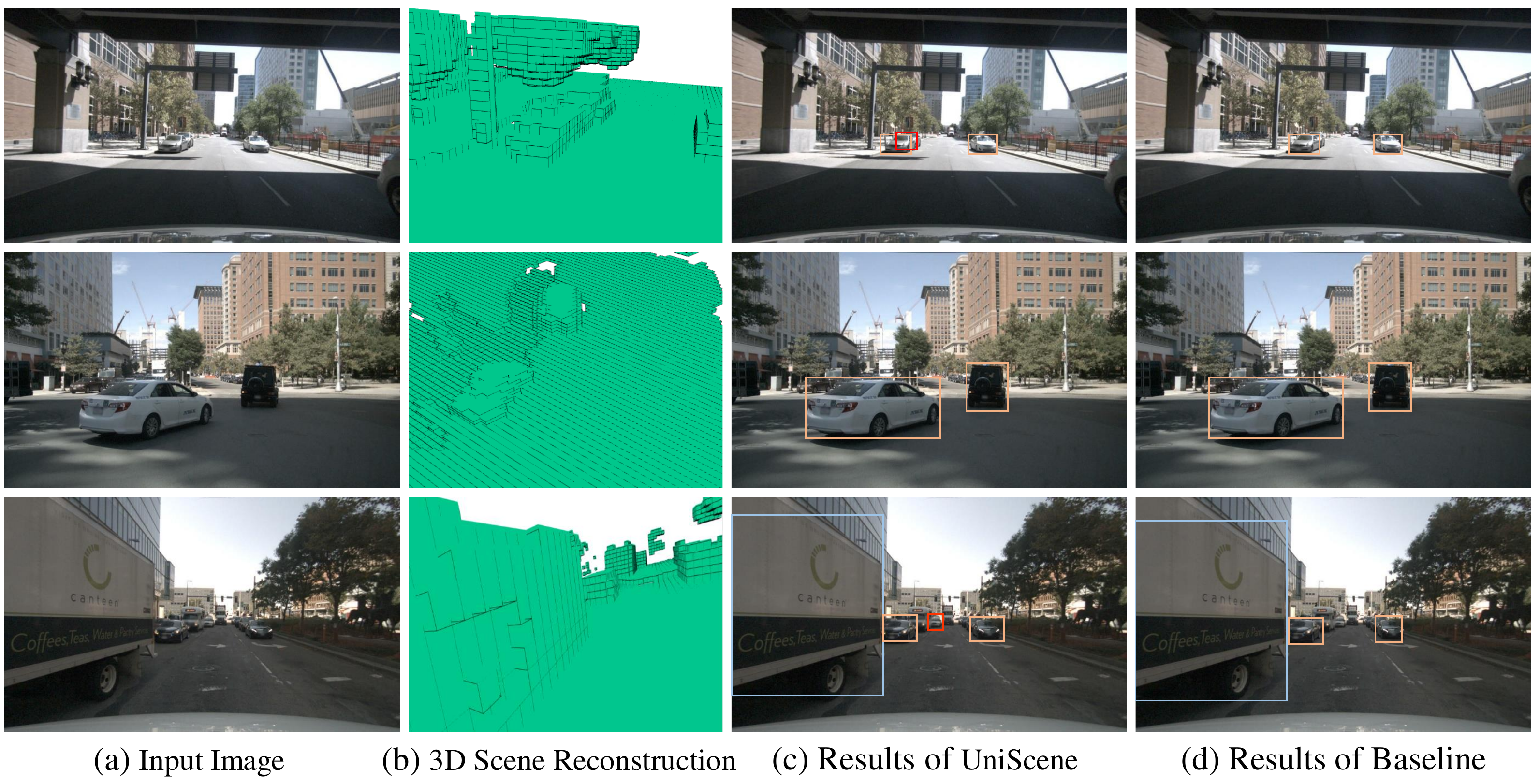} 
	\caption{Visualization of 3D scene reconstruction and results of 3D object detection.}
	\label{fig:show}
\end{figure*}

\begin{table}[t]
	\centering
	\caption{Impacts of multi-frame fusion.}
	\setlength{\tabcolsep}{5.9mm}
	{
		\begin{tabular}{c|c|c|cc}
			\toprule
			\textbf{Frames}&\textbf{mAP}$\uparrow$&\textbf{NDS}$\uparrow$&\textbf{mATE}$\downarrow$\\
			\midrule
			Baseline &0.416&0.517&0.673 \\
			1 &0.417&0.523&0.661 \\
			3 &\textbf{0.438}&\textbf{0.534}&\textbf{0.656}\\ 
			5 &0.429&0.525&0.659  \\
			\bottomrule
		\end{tabular}
	}
	\label{tab:fusion}
\end{table}

\subsection{Ablation Studies}
In this section, we perform thorough ablation experiments with BEVFormer~\cite{bevformer} on nuScenes validation set.
\subsubsection{Data-efficient Learner}

Fine-tuning models using limited labeled data is made possible through pre-training. In order to evaluate the data efficiency of UniScene, we conducted experiments using varying amounts of labeled data for fine-tuning. BEVFormer~\cite{bevformer} was utilized as the backbone and assessed the detection performance of the model on the nuScenes validation set. The results, as depicted in Figure~\ref{fig:efficient}, demonstrate that when BEVFormer is trained with 75\% of the labeled data, it achieves the same performance as it trained on the complete dataset. Moreover, even with only 25\% of the samples available for fine-tuning, our UniScene model outperforms BEVFormer by 1\% in mAP, highlighting its remarkable data efficiency and its potential to reduce the reliance on expensive human-annotated 3D data.

\subsubsection{Multi-frame Fusion}

We conducted an analysis of the influence of the number of fused LiDAR frames on the pre-trained model. As more frames were fused, the density of the point clouds increased. Our comparison included single-frame fusion, 3-frame fusion and 5-frame fusion (including the corresponding non-key frames) and the results are presented in Table~\ref{tab:fusion}. It is clear that the model's accuracy initially improved with an increasing number of fused point clouds but started to decline afterwards. This finding suggests that fusing multiple frames of point clouds can enhance the effectiveness of the pre-trained model. However, it is important to note that excessive fusion of frames introduces uncertainty due to the presence of dynamic objects. This uncertainty can lead to errors in the fusion process and subsequently lower the accuracy of the model.

\subsubsection{Explicit Supervision}
\begin{table}[t]
	\centering
	\caption{Semantic occupancy supervision.}
	\setlength{\tabcolsep}{5.3mm}
	{
		\begin{tabular}{c|c|c|cc}
			\toprule
			\textbf{Methods}&\textbf{mAP}$\uparrow$&\textbf{NDS}$\uparrow$&\textbf{mATE}$\downarrow$\\
			\midrule
			BEVFormer~\cite{bevformer} &0.416&0.517&0.673 \\
			UniScene &0.438&0.534&0.656  \\
			\midrule
			Supervision &\textbf{0.445}&\textbf{0.544}&\textbf{0.648} \\
			\bottomrule
	\end{tabular}}
	\label{tab:supervision}
\end{table}

Labeled 3D data can be utilized to handle dynamic objects separately during the point cloud fusion process, resulting in more precise occupancy grid ground truth for multi-frame fusion. Subsequently, we examined the impact of explicit occupancy grid prediction on the model's performance. The results in Table~\ref{tab:supervision} demonstrate that incorporating explicit supervision leads to a notable improvement of 3\% in mAP and NDS compared to BEVFormer~\cite{bevformer}. Furthermore, when compared to unlabeled multi-frame fused point cloud pre-training, there is a 1\% increase in mAP. These findings highlight the potential of leveraging labeled data for explicit occupancy prediction supervision. Moreover, they further support the proposition that occupancy prediction enables the model to learn the data distribution of the entire 3D scene, thereby enhancing the accuracy of downstream tasks.

\subsection{Qualitative Evaluation}
As shown in Figure~\ref{fig:show}, we present several reconstructed 3D scenes. It can be observed that UniScene can reconstruct complete 3D objects, including handling occlusions and distant objects effectively. Compared to baseline pre-training models that estimate depth on object surfaces, UniScene can provide comprehensive 3D geometric information, thereby enhancing the performance of downstream tasks.

\textbf{Limitations.} Although our multi-camera unified pre-training approach has demonstrated promising results, there are several limitations to consider:
(1) The 3D convolutions in the decoder limit its applicability to tasks requiring high-resolution occupancy reconstruction. We will explore the cascade refine strategy. 
(2) UniScenes is conducted by reconstructing 3D static scenes, with models lacking predictive capabilities. In the future, pre-training will focus on 4D World Models~\cite{ha2018world,hafner2020mastering}.

\section{Conclusion}
We first define the task of multi-camera unified pre-training and propose the unified pre-training algorithm that enables the model to grasp geometric priors of the surrounding world. Pre-training via 3D scene reconstruction with unlabeled image-LiDAR pairs offers promising opportunities for reducing the dependency on annotated 3D data and establishing a foundational model for autonomous driving. Future work should focus on addressing the limitations mentioned and further improving the performance and applicability of our approach in real-world autonomous driving scenarios.

\bibliographystyle{IEEEtran}
\bibliography{references}

\end{document}

%% file: shortcuts.tex
\usepackage{color}
\definecolor{red}{rgb}{1.00,0.20,0.20}
\definecolor{blue}{rgb}{0.20,0.20,1.00}
\definecolor{green}{rgb}{0.00,1.00,0.00}

\usepackage{array}
\newcolumntype{L}[1]{>{\raggedright\let\newline\\\arraybackslash\hspace{0pt}}m{#1}}
\newcolumntype{C}[1]{>{\centering\let\newline\\\arraybackslash\hspace{0pt}}m{#1}}
\newcolumntype{R}[1]{>{\raggedleft\let\newline\\\arraybackslash\hspace{0pt}}m{#1}}


%% file: acronyms.tex
\ifdefined\UseShortAcronyms

\else

\fi

%% file: root.bbl
\begin{thebibliography}{10}
\providecommand{\url}[1]{#1}
\csname url@samestyle\endcsname
\providecommand{\newblock}{\relax}
\providecommand{\bibinfo}[2]{#2}
\providecommand{\BIBentrySTDinterwordspacing}{\spaceskip=0pt\relax}
\providecommand{\BIBentryALTinterwordstretchfactor}{4}
\providecommand{\BIBentryALTinterwordspacing}{\spaceskip=\fontdimen2\font plus
\BIBentryALTinterwordstretchfactor\fontdimen3\font minus
  \fontdimen4\font\relax}
\providecommand{\BIBforeignlanguage}[2]{{%
\expandafter\ifx\csname l@#1\endcsname\relax
\typeout{** WARNING: IEEEtran.bst: No hyphenation pattern has been}%
\typeout{** loaded for the language `#1'. Using the pattern for}%
\typeout{** the default language instead.}%
\else
\language=\csname l@#1\endcsname
\fi
#2}}
\providecommand{\BIBdecl}{\relax}
\BIBdecl

\bibitem{survey1}
Y.~Ma, T.~Wang, X.~Bai, H.~Yang, Y.~Hou, Y.~Wang, Y.~Qiao, R.~Yang, D.~Manocha,
  and X.~Zhu, ``Vision-centric bev perception: A survey,'' \emph{arXiv preprint
  arXiv:2208.02797}, 2022.

\bibitem{survey2}
H.~Li, C.~Sima, J.~Dai, W.~Wang, L.~Lu, H.~Wang, E.~Xie, Z.~Li, H.~Deng,
  H.~Tian \emph{et~al.}, ``Delving into the devils of bird's-eye-view
  perception: A review, evaluation and recipe,'' \emph{arXiv preprint
  arXiv:2209.05324}, 2022.

\bibitem{bevfusion}
T.~Liang, H.~Xie, K.~Yu, Z.~Xia, Z.~Lin, Y.~Wang, T.~Tang, B.~Wang, and
  Z.~Tang, ``Bevfusion: A simple and robust lidar-camera fusion framework,''
  \emph{arXiv preprint arXiv:2205.13790}, 2022.

\bibitem{chen2023futr3d}
X.~Chen, T.~Zhang, Y.~Wang, Y.~Wang, and H.~Zhao, ``Futr3d: A unified sensor
  fusion framework for 3d detection,'' in \emph{Proceedings of the IEEE/CVF
  Conference on Computer Vision and Pattern Recognition}, 2023, pp. 172--181.

\bibitem{li2022unifying}
Y.~Li, Y.~Chen, X.~Qi, Z.~Li, J.~Sun, and J.~Jia, ``Unifying voxel-based
  representation with transformer for 3d object detection,'' \emph{arXiv
  preprint arXiv:2206.00630}, 2022.

\bibitem{detr3d}
Y.~Wang, V.~C. Guizilini, T.~Zhang, Y.~Wang, H.~Zhao, and J.~Solomon, ``Detr3d:
  3d object detection from multi-view images via 3d-to-2d queries,'' in
  \emph{Conference on Robot Learning}.\hskip 1em plus 0.5em minus 0.4em\relax
  PMLR, 2022, pp. 180--191.

\bibitem{bevformer}
Z.~Li, W.~Wang, H.~Li, E.~Xie, C.~Sima, T.~Lu, Q.~Yu, and J.~Dai, ``Bevformer:
  Learning bird's-eye-view representation from multi-camera images via
  spatiotemporal transformers,'' \emph{arXiv preprint arXiv:2203.17270}, 2022.

\bibitem{bevdet}
J.~Huang, G.~Huang, Z.~Zhu, and D.~Du, ``Bevdet: High-performance multi-camera
  3d object detection in bird-eye-view,'' \emph{arXiv preprint
  arXiv:2112.11790}, 2021.

\bibitem{bevdepth}
Y.~Li, Z.~Ge, G.~Yu, J.~Yang, Z.~Wang, Y.~Shi, J.~Sun, and Z.~Li, ``Bevdepth:
  Acquisition of reliable depth for multi-view 3d object detection,''
  \emph{arXiv preprint arXiv:2206.10092}, 2022.

\bibitem{imagenet}
J.~Deng, W.~Dong, R.~Socher, L.-J. Li, K.~Li, and L.~Fei-Fei, ``Imagenet: A
  large-scale hierarchical image database,'' in \emph{2009 IEEE conference on
  computer vision and pattern recognition}.\hskip 1em plus 0.5em minus
  0.4em\relax Ieee, 2009, pp. 248--255.

\bibitem{dd3d}
D.~Park, R.~Ambrus, V.~Guizilini, J.~Li, and A.~Gaidon, ``Is pseudo-lidar
  needed for monocular 3d object detection?'' in \emph{Proceedings of the
  IEEE/CVF International Conference on Computer Vision}, 2021, pp. 3142--3152.

\bibitem{occnet}
W.~Tong, C.~Sima, T.~Wang, S.~Wu, H.~Deng, L.~Chen, Y.~Gu, L.~Lu, P.~Luo,
  D.~Lin \emph{et~al.}, ``Scene as occupancy,'' \emph{arXiv preprint
  arXiv:2306.02851}, 2023.

\bibitem{occ_survey}
Y.~Shi, K.~Jiang, J.~Li, J.~Wen, Z.~Qian, M.~Yang, K.~Wang, and D.~Yang,
  ``Grid-centric traffic scenario perception for autonomous driving: A
  comprehensive review,'' \emph{arXiv preprint arXiv:2303.01212}, 2023.

\bibitem{li2023voxformer}
Y.~Li, Z.~Yu, C.~Choy, C.~Xiao, J.~M. Alvarez, S.~Fidler, C.~Feng, and
  A.~Anandkumar, ``Voxformer: Sparse voxel transformer for camera-based 3d
  semantic scene completion,'' in \emph{Proceedings of the IEEE/CVF Conference
  on Computer Vision and Pattern Recognition}, 2023, pp. 9087--9098.

\bibitem{lss}
J.~Philion and S.~Fidler, ``Lift, splat, shoot: Encoding images from arbitrary
  camera rigs by implicitly unprojecting to 3d,'' in \emph{European Conference
  on Computer Vision}.\hskip 1em plus 0.5em minus 0.4em\relax Springer, 2020,
  pp. 194--210.

\bibitem{nuscenes}
H.~Caesar, V.~Bankiti, A.~H. Lang, S.~Vora, V.~E. Liong, Q.~Xu, A.~Krishnan,
  Y.~Pan, G.~Baldan, and O.~Beijbom, ``nuscenes: A multimodal dataset for
  autonomous driving,'' in \emph{Proceedings of the IEEE/CVF conference on
  computer vision and pattern recognition}, 2020, pp. 11\,621--11\,631.

\bibitem{zhang2022beverse}
Y.~Zhang, Z.~Zhu, W.~Zheng, J.~Huang, G.~Huang, J.~Zhou, and J.~Lu, ``Beverse:
  Unified perception and prediction in birds-eye-view for vision-centric
  autonomous driving,'' \emph{arXiv preprint arXiv:2205.09743}, 2022.

\bibitem{videobev}
C.~Han, J.~Sun, Z.~Ge, J.~Yang, R.~Dong, H.~Zhou, W.~Mao, Y.~Peng, and
  X.~Zhang, ``Exploring recurrent long-term temporal fusion for multi-view 3d
  perception,'' \emph{arXiv preprint arXiv:2303.05970}, 2023.

\bibitem{sts}
Z.~Wang, C.~Min, Z.~Ge, Y.~Li, Z.~Li, H.~Yang, and D.~Huang, ``Sts:
  Surround-view temporal stereo for multi-view 3d detection,'' \emph{arXiv
  preprint arXiv:2208.10145}, 2022.

\bibitem{solofusion}
J.~Park, C.~Xu, S.~Yang, K.~Keutzer, K.~Kitani, M.~Tomizuka, and W.~Zhan,
  ``Time will tell: New outlooks and a baseline for temporal multi-view 3d
  object detection,'' \emph{arXiv preprint arXiv:2210.02443}, 2022.

\bibitem{bevstereo}
Y.~Li, H.~Bao, Z.~Ge, J.~Yang, J.~Sun, and Z.~Li, ``Bevstereo: Enhancing depth
  estimation in multi-view 3d object detection with dynamic temporal stereo,''
  \emph{arXiv preprint arXiv:2209.10248}, 2022.

\bibitem{petr}
Y.~Liu, T.~Wang, X.~Zhang, and J.~Sun, ``Petr: Position embedding
  transformation for multi-view 3d object detection,'' \emph{arXiv preprint
  arXiv:2203.05625}, 2022.

\bibitem{uniad}
Y.~Hu, J.~Yang, L.~Chen, K.~Li, C.~Sima, X.~Zhu, S.~Chai, S.~Du, T.~Lin,
  W.~Wang, L.~Lu, X.~Jia, Q.~Liu, J.~Dai, Y.~Qiao, and H.~Li,
  ``Planning-oriented autonomous driving,'' in \emph{Proceedings of the
  IEEE/CVF Conference on Computer Vision and Pattern Recognition}, 2023.

\bibitem{deepcluster}
M.~Caron, P.~Bojanowski, A.~Joulin, and M.~Douze, ``Deep clustering for
  unsupervised learning of visual features,'' in \emph{Proceedings of the
  European conference on computer vision (ECCV)}, 2018, pp. 132--149.

\bibitem{swav}
M.~Caron, I.~Misra, J.~Mairal, P.~Goyal, P.~Bojanowski, and A.~Joulin,
  ``Unsupervised learning of visual features by contrasting cluster
  assignments,'' \emph{Advances in Neural Information Processing Systems},
  vol.~33, pp. 9912--9924, 2020.

\bibitem{moco}
K.~He, H.~Fan, Y.~Wu, S.~Xie, and R.~Girshick, ``Momentum contrast for
  unsupervised visual representation learning,'' in \emph{Proceedings of the
  IEEE/CVF conference on computer vision and pattern recognition}, 2020, pp.
  9729--9738.

\bibitem{byol}
J.-B. Grill, F.~Strub, F.~Altch{\'e}, C.~Tallec, P.~Richemond, E.~Buchatskaya,
  C.~Doersch, B.~Avila~Pires, Z.~Guo, M.~Gheshlaghi~Azar \emph{et~al.},
  ``Bootstrap your own latent-a new approach to self-supervised learning,''
  \emph{Advances in neural information processing systems}, vol.~33, pp.
  21\,271--21\,284, 2020.

\bibitem{mae}
K.~He, X.~Chen, S.~Xie, Y.~Li, P.~Doll{\'a}r, and R.~Girshick, ``Masked
  autoencoders are scalable vision learners,'' in \emph{Proceedings of the
  IEEE/CVF Conference on Computer Vision and Pattern Recognition}, 2022, pp.
  16\,000--16\,009.

\bibitem{beit}
H.~Bao, L.~Dong, S.~Piao, and F.~Wei, ``Beit: Bert pre-training of image
  transformers,'' \emph{arXiv preprint arXiv:2106.08254}, 2021.

\bibitem{voxel-mae}
C.~Min, D.~Zhao, L.~Xiao, Y.~Nie, and B.~Dai, ``Voxel-mae: Masked autoencoders
  for pre-training large-scale point clouds,'' \emph{arXiv preprint
  arXiv:2206.09900}, 2022.

\bibitem{also}
A.~Boulch, C.~Sautier, B.~Michele, G.~Puy, and R.~Marlet, ``Also: Automotive
  lidar self-supervision by occupancy estimation,'' \emph{arXiv preprint
  arXiv:2212.05867}, 2022.

\bibitem{iscc}
Z.~Zhang, M.~Bai, and E.~Li, ``Implicit surface contrastive clustering for
  lidar point clouds,'' in \emph{Proceedings of the IEEE/CVF Conference on
  Computer Vision and Pattern Recognition}, 2023, pp. 21\,716--21\,725.

\bibitem{resnet}
K.~He, X.~Zhang, S.~Ren, and J.~Sun, ``Deep residual learning for image
  recognition,'' in \emph{Proceedings of the IEEE conference on computer vision
  and pattern recognition}, 2016, pp. 770--778.

\bibitem{second}
Y.~Yan, Y.~Mao, and B.~Li, ``Second: Sparsely embedded convolutional
  detection,'' \emph{Sensors}, vol.~18, no.~10, p. 3337, 2018.

\bibitem{pv_rcnn}
S.~Shi, C.~Guo, L.~Jiang, Z.~Wang, J.~Shi, X.~Wang, and H.~Li, ``Pv-rcnn:
  Point-voxel feature set abstraction for 3d object detection,'' in
  \emph{Proceedings of the IEEE/CVF Conference on Computer Vision and Pattern
  Recognition}, 2020, pp. 10\,529--10\,538.

\bibitem{centerpoint}
T.~Yin, X.~Zhou, and P.~Krahenbuhl, ``Center-based 3d object detection and
  tracking,'' in \emph{Proceedings of the IEEE/CVF conference on computer
  vision and pattern recognition}, 2021, pp. 11\,784--11\,793.

\bibitem{cylinder3d}
X.~Zhu, H.~Zhou, T.~Wang, F.~Hong, Y.~Ma, W.~Li, H.~Li, and D.~Lin,
  ``Cylindrical and asymmetrical 3d convolution networks for lidar
  segmentation,'' in \emph{Proceedings of the IEEE/CVF conference on computer
  vision and pattern recognition}, 2021, pp. 9939--9948.

\bibitem{tpvformer}
Y.~Huang, W.~Zheng, Y.~Zhang, J.~Zhou, and J.~Lu, ``Tri-perspective view for
  vision-based 3d semantic occupancy prediction,'' \emph{arXiv preprint
  arXiv:2302.07817}, 2023.

\bibitem{openoccupancy}
X.~Wang, Z.~Zhu, W.~Xu, Y.~Zhang, Y.~Wei, X.~Chi, Y.~Ye, D.~Du, J.~Lu, and
  X.~Wang, ``Openoccupancy: A large scale benchmark for surrounding semantic
  occupancy perception,'' \emph{arXiv preprint arXiv:2303.03991}, 2023.

\bibitem{occ3d}
X.~Tian, T.~Jiang, L.~Yun, Y.~Wang, Y.~Wang, and H.~Zhao, ``Occ3d: A
  large-scale 3d occupancy prediction benchmark for autonomous driving,''
  \emph{arXiv preprint arXiv:2304.14365}, 2023.

\bibitem{gan2023simple}
W.~Gan, N.~Mo, H.~Xu, and N.~Yokoya, ``A simple attempt for 3d occupancy
  estimation in autonomous driving,'' \emph{arXiv preprint arXiv:2303.10076},
  2023.

\bibitem{bevdistill}
Z.~Chen, Z.~Li, S.~Zhang, L.~Fang, Q.~Jiang, and F.~Zhao, ``Bevdistill:
  Cross-modal bev distillation for multi-view 3d object detection,''
  \emph{arXiv preprint arXiv:2211.09386}, 2022.

\bibitem{tigbev}
P.~Huang, L.~Liu, R.~Zhang, S.~Zhang, X.~Xu, B.~Wang, and G.~Liu, ``Tig-bev:
  Multi-view bev 3d object detection via target inner-geometry learning,''
  \emph{arXiv preprint arXiv:2212.13979}, 2022.

\bibitem{geomim}
J.~Liu, T.~Wang, B.~Liu, Q.~Zhang, Y.~Liu, and H.~Li, ``Towards better 3d
  knowledge transfer via masked image modeling for multi-view 3d
  understanding,'' \emph{arXiv preprint arXiv:2303.11325}, 2023.

\bibitem{nerf}
B.~Mildenhall, P.~P. Srinivasan, M.~Tancik, J.~T. Barron, R.~Ramamoorthi, and
  R.~Ng, ``Nerf: Representing scenes as neural radiance fields for view
  synthesis,'' \emph{Communications of the ACM}, vol.~65, no.~1, pp. 99--106,
  2021.

\bibitem{nerf2}
J.~T. Barron, B.~Mildenhall, M.~Tancik, P.~Hedman, R.~Martin-Brualla, and P.~P.
  Srinivasan, ``Mip-nerf: A multiscale representation for anti-aliasing neural
  radiance fields,'' in \emph{Proceedings of the IEEE/CVF International
  Conference on Computer Vision}, 2021, pp. 5855--5864.

\bibitem{mvs}
Q.~Zhu, C.~Min, Z.~Wei, Y.~Chen, and G.~Wang, ``Deep learning for multi-view
  stereo via plane sweep: A survey,'' \emph{arXiv preprint arXiv:2106.15328},
  2021.

\bibitem{mvsnet}
Y.~Yao, Z.~Luo, S.~Li, T.~Fang, and L.~Quan, ``Mvsnet: Depth inference for
  unstructured multi-view stereo,'' in \emph{Proceedings of the European
  conference on computer vision (ECCV)}, 2018, pp. 767--783.

\bibitem{aa-rmvsnet}
Z.~Wei, Q.~Zhu, C.~Min, Y.~Chen, and G.~Wang, ``Aa-rmvsnet: Adaptive
  aggregation recurrent multi-view stereo network,'' in \emph{Proceedings of
  the IEEE/CVF International Conference on Computer Vision}, 2021, pp.
  6187--6196.

\bibitem{bi}
Z.~Wei, Q.~Zhu, C.~Min, and G.~Wang, ``Bidirectional hybrid lstm based
  recurrent neural network for multi-view stereo,'' \emph{IEEE Transactions on
  Visualization and Computer Graphics}, 2022.

\bibitem{tesla}
.~T.~A. Day, ``[online],''
  \url{http://https://www.youtube.com/watch?v=jPCV4GKX9Dw}.

\bibitem{fcos3d}
T.~Wang, X.~Zhu, J.~Pang, and D.~Lin, ``{FCOS3D: Fully} convolutional one-stage
  monocular 3d object detection,'' in \emph{Proceedings of the IEEE/CVF
  International Conference on Computer Vision (ICCV) Workshops}, 2021.

\bibitem{occ}
\BIBentryALTinterwordspacing
``Cvpr 2023 occupancy prediction challenge,'' 2023. [Online]. Available:
  \url{https://github.com/CVPR2023-3D-Occupancy-Prediction/CVPR2023-3D-Occupancy-Prediction}
\BIBentrySTDinterwordspacing

\bibitem{swin}
Z.~Liu, Y.~Lin, Y.~Cao, H.~Hu, Y.~Wei, Z.~Zhang, S.~Lin, and B.~Guo, ``Swin
  transformer: Hierarchical vision transformer using shifted windows,'' in
  \emph{Proceedings of the IEEE/CVF international conference on computer
  vision}, 2021, pp. 10\,012--10\,022.

\bibitem{waymo}
P.~Sun, H.~Kretzschmar, X.~Dotiwalla, A.~Chouard, V.~Patnaik, P.~Tsui, J.~Guo,
  Y.~Zhou, Y.~Chai, B.~Caine \emph{et~al.}, ``Scalability in perception for
  autonomous driving: Waymo open dataset,'' in \emph{Proceedings of the
  IEEE/CVF conference on computer vision and pattern recognition}, 2020, pp.
  2446--2454.

\bibitem{kitti}
A.~Geiger, P.~Lenz, and R.~Urtasun, ``Are we ready for autonomous driving? the
  kitti vision benchmark suite,'' in \emph{2012 IEEE conference on computer
  vision and pattern recognition}.\hskip 1em plus 0.5em minus 0.4em\relax IEEE,
  2012, pp. 3354--3361.

\bibitem{ha2018world}
D.~Ha and J.~Schmidhuber, ``World models,'' \emph{arXiv preprint
  arXiv:1803.10122}, 2018.

\bibitem{hafner2020mastering}
D.~Hafner, T.~Lillicrap, M.~Norouzi, and J.~Ba, ``Mastering atari with discrete
  world models,'' \emph{arXiv preprint arXiv:2010.02193}, 2020.

\end{thebibliography}
